\documentclass{article}
\usepackage{spconf,amsmath,epsfig}
\usepackage{multirow}
\usepackage{color}
\definecolor{BlueColor}{rgb}{0.0, 0.0, 1.0}
\definecolor{RedColor}{rgb}{1.0, 0.0, 0.0}

\let\OLDthebibliography\thebibliography
\renewcommand\thebibliography[1]{
  \OLDthebibliography{#1}
  \setlength{\parskip}{0pt}
  \setlength{\itemsep}{0pt plus 0.3ex}
}

\begin{document}\sloppy

\def\x{{\mathbf x}}
\def\L{{\cal L}}

\title{Dynamic Spatial-temporal Hypergraph Convolutional Network for Skeleton-based Action Recognition}
%

\name{Shengqin Wang, Yongji Zhang, Hong Qi, Minghao Zhao, Yu Jiang}
\address{}

\maketitle
\begin{abstract}

Skeleton-based action recognition relies on the extraction of spatial-temporal topological information. Hypergraphs can establish prior unnatural dependencies for the skeleton. However, the existing methods only focus on the construction of spatial topology and ignore the time-point dependence. This paper proposes a dynamic spatial-temporal hypergraph convolutional network (DST-HCN) to capture spatial-temporal information for skeleton-based action recognition. DST-HCN introduces a time-point hypergraph (TPH) to learn relationships at time points. With multiple spatial static hypergraphs and dynamic TPH, our network can learn more complete spatial-temporal features. In addition, we use the high-order information fusion module (HIF) to fuse spatial-temporal information synchronously. Extensive experiments on NTU RGB+D, NTU RGB+D 120, and NW-UCLA datasets show that our model achieves state-of-the-art, especially compared with hypergraph methods.
\end{abstract}


\section{Introduction}
\label{sec:intro}

Action recognition is one of the important tasks in computer vision, widely used in human-robot interaction, video understanding and intelligent detection. In recent years, skeleton-based action recognition has attracted increasing attention due to its computational efficiency and robustness to background or environmental changes.
\begin{figure}[t] 
\centering 
\includegraphics[height=1.2\linewidth,width=0.8\linewidth]{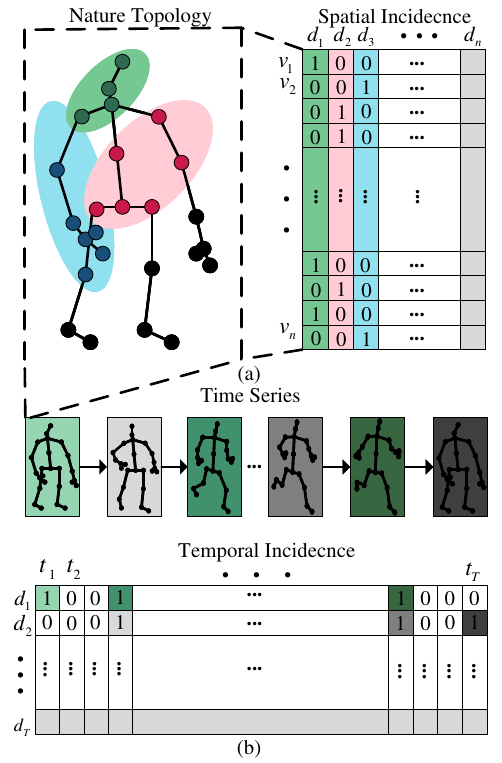} 
\caption{Skeleton representation based on hypergraph. (a) is the spatial topological hypergraph generated by k-means method and different colors indicate different hyperedges. (b) is the time-point hypergraph generated by k-NN and the depth of the colour indicates the importance of time.}
\label{fig01} 
\end{figure}

\begin{figure}[t]
\centering 
\includegraphics[height=1\linewidth,width=0.9\linewidth]{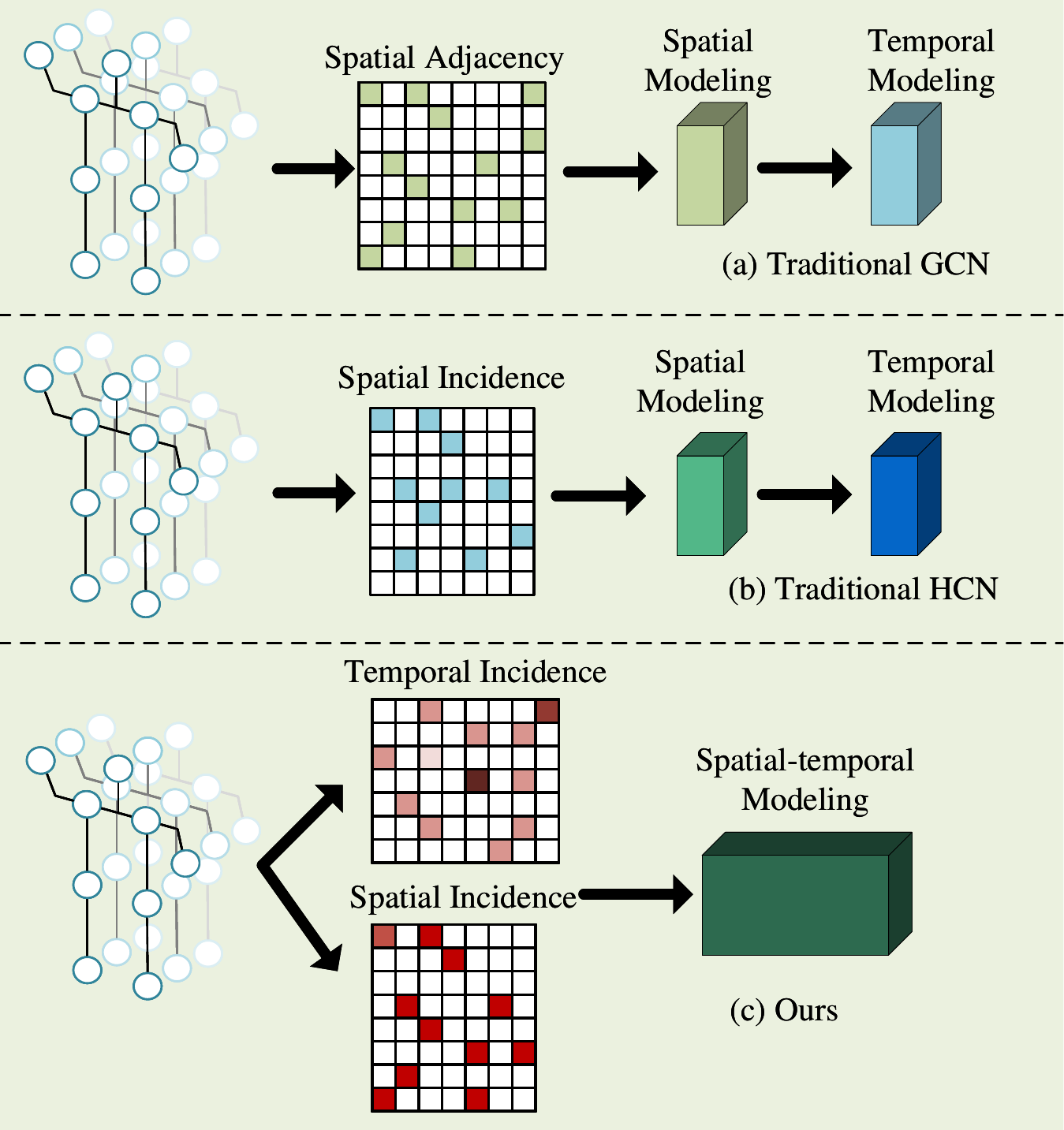} 
\caption{Comparison of our proposed framework with other skeleton recognition methods. (a) is graph convolution method. (b) is hypergraph convolution method. (c) is our proposed method.}
\label{fig002}
\end{figure}

Graph convolution methods are widely used to handle the spatial-temporal topology of the human skeleton. ST-GCN~\cite{DBLP:conf/aaai/YanXL18} proposed a joint segmentation strategy to delineate the joint points. However, using the human body's static segmentation strategy is insufficient to adapt to the numerous actions. Therefore, Chen et al.~\cite{2021Channel} proposed CTR-GCN that generates new topologies based on sample features dynamically. But the spatial topology used in these methods~\cite{DBLP:conf/aaai/YanXL18,2020Semantics,2021Channel} hardly reflects the multiple joints of unnatural dependencies and does not consider the issue of time-point dependencies. The emergence of hypergraph convolution-based approach~\cite{DBLP:journals/tip/HaoLGJY21,wei2021dynamic,zhu2022selective} considers unnatural dependencies beyond the natural topology. While these methods also ignore the temporal dependencies between actions to a certain extent. In addition, both hypergraph and graph convolution methods~\cite{2020Semantics,DBLP:conf/icmcs/ChenXJS21,2021Channel} use a strategy of learning spatial-temporal separately. This results in the spatial-temporal parallel information of action sequences being ignored.
To solve the above problem, we construct a dynamic spatial-temporal hypergraph convolutional network (DST-HCN) that consists of dynamic spatial-temporal hypergraph convolution (DST-HC) and high-order information fusion (HIF). We use the K-NN method in DST-HC to generate dynamic time-point hypergraph (TPH) based on sample features. For the spatial topology, we construct three hypergraph topologies using methods such as k-means. The overview is clearly shown in Fig.~\ref{fig01}. TPH and constructed spatial hypergraph topology are passed through a convolution operation to obtain high-order spatial-temporal feature information. In HIF, action modeling is done by aggregating and concatenating the high-order information through multi-feature channel operations. As shown in Fig.~\ref{fig002}, we introduce TPH to solve the adjacent topological similarity problem. In addition, we use spatial-temporal parallel learning to fuse features efficiently.

In summary, our contributions are mainly in the following areas:
\begin{itemize}

\item We propose the DST-HCN that captures important spatial-temporal feature information based on the constructed multiple hypergraphs. And an effective fusion is done through parallel spatial-temporal modeling to finalize the action recognition task.


\item We introduce the time-point hypergraph.  By combining spatial hypergraphs, we construct completable spatial-temporal dependencies.

\item Extensive experiments show that DST-HCN achieves state-of-the-art on the NTU RGB+D, NTU RGB+D 120, and NW-UCLA datasets.
\end{itemize}

\section{RELATED WORK}

\subsection{Skeleton-based Action Recognition}
CNNs have achieved remarkable results in processing euclidean data like images and RNNs have achieved significant advantages in processing time series.  So the research began to emerge with CNN and RNN methods~\cite{2016NTU,DBLP:conf/icmcs/ZhangSZ19,xu2022topology}. The advent of graph convolutional neural networks (GCNs) have significantly improved the processing of non-Euclidean data. Yan et al.~\cite{DBLP:conf/aaai/YanXL18} proposed the ST-GCN model, which can better handle the topological relationship of skeletons to achieve more effectiveness. Chen et al.~\cite{2021Channel} proposed the CTR-GCN method, which can dynamically learn joint information on space to achieve channel topology modeling. Song et al.~\cite{2021Constructing} proposed a family of efficient GCN baselines with high accuracies and small amounts of trainable parameters. This methods use a strategy of spatial-temporal separate learning in model configuration. 
\subsection{Hypergraph Neural Networks}
The graph structure used in traditional graph convolutional neural networks focuses on establishing a correspondence between two objects to form a set. Hyperedges can connect arbitrary objects to form hypergraphs to obtain high-order dependencies. Feng et al.~\cite{DBLP:conf/aaai/FengYZJG19} proposed hypergraph neural network (HGNN) to learn the hidden layer representation considering the high-order data structure. Jo et al.~\cite{DBLP:conf/nips/JoBLKKH21} proposed a pairwise hypergraph transformation method that makes the network focus on the feature information of the edges. Jiang et al.~\cite{jiang2019dynamic} proposed a dynamic hypergraph network (DHNN) to update the hypergraph structure dynamically, etc. In this paper, we use the generic hypergraph convolution described with HGNN~\cite{DBLP:conf/aaai/FengYZJG19}.

\section{METHOD}

\begin{figure*}[htbp]
\centerline{\includegraphics[width=16cm,height = 6cm]{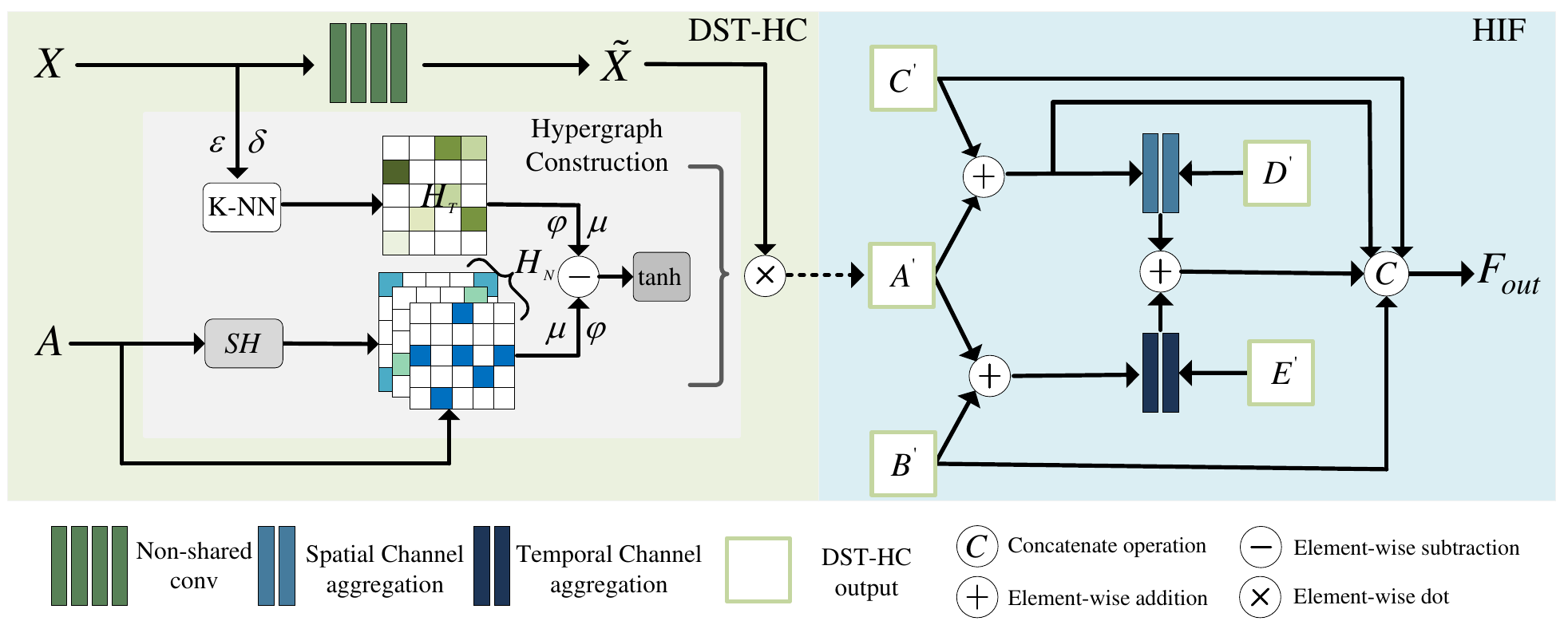}}
\caption{The framework for our dynamic spatial-temporal hypergraph neural network. DST-HC constructs multiple hypergraphs and then obtains high-order spatial-temporal feature information by convolution operations. The HIF aims at performing spatial-temporal parallel modeling in the channel dimension.}
\label{fig02}
\end{figure*}
\subsection{Preliminaries}
\noindent
{\bf Graph Convolutional Network.} A skeleton sequence consists of T single-frame skeletal maps, which is represented by $G=(V, E)$~\cite{DBLP:conf/aaai/YanXL18}, where $V$ is the set of joints, and $E$ is the set of skeletal edges. The skeleton graph is denoted by $A\in R^{N\times N}$, which is the adjacency matrix about the joints. A conventional graph convolution method based on skeletal data is:

\begin{align}
X_{i+1}=\sum _{A{i}\in A_{k}}X_{i}\cdot  \widetilde{A_{i}}\cdot M_{i},
\end{align}

where $X_{i}\in R^{C_{in}\times T\times V}$ denotes the input features of the i-th layer of the network, $C_{in}$ is the number of channels, $\widetilde{A_{i}}$ denotes the adjacency matrix of the normalised spatial segmentation method, $A_{k}$ denotes spatial skeletal segmentation method~\cite{DBLP:conf/aaai/YanXL18}, and $ M_{i}$ denotes the learnable weight features.


\subsection{Dynamic Spatial-temporal Hypergraph Convolution (DST-HC)}
\label{BB}

\noindent
{\bf Hypergraph Construction.} Unlike the graph convolution method, the hypergraph structure is represented by $g=(v,d,w)$, where $v$ denotes the number of joints, $d$ is the set of hyperedges represented by the incidence matrix, and $w$ is represented as the set of weights of hyperedges. For the incidence matrix $H\in R^{V\times N} $, $V$ is the set of joints and $N$ denotes the number of hyperedges. For each hyperedge: $h(v,e)=1$ means the joint belongs to this hyperedges, and $h(v,e)=0$ means there is no node in this hyperedges. 

In the dynamic time-point hypergraph construction method, we use the k-NN method to construct the features after linear transformation and dimensionality reduction to reduce the computational cost. As shown in Fig.~\ref{fig02}, the hypergraph $H_{T}\in R^{T\times T}$ is formulated as:
\begin{align}
H_{T} =TH(\delta(\varepsilon(X) ,k),
\end{align}
where $k$ denotes the number of neighbors of the each hyperedge, $X\in R^{C_{in}\times T\times V} $ denotes the input features, $\varepsilon(\cdot)$ is the linear transformation dimensionality reduction function, $\delta(\cdot)$ is the dimensionality transformation function, and $TH(\cdot)$ denotes the k-NN function.

We use k-NN, k-means, and the combination of centrifugal and centripetal joints~\cite{DBLP:conf/aaai/YanXL18} to construct the spatial hypergraph matrix, as shown in Fig.~\ref{fig02}. The spatial hypergraph method is formulated as:
\begin{align}
H_{N} =SH(A),
\end{align} 
where $SH(\cdot)$ are the three methods mentioned above, and $A$ is the topological adjacency matrix.

Further, we construct two spatial-temporal dependencies $H_{ST}\in R^{V\times T}, H_{TS}\in R^{T\times V}$ based on the above hypergraph. As shown in the following equation:
\begin{align}
H_{ST}/H_{TS}=\Phi ((H_{N},H_{T})/(H_{T},H_{N})),
\end{align}
where $\Phi(a,b)=tanh (\mu(a)-\varphi(b))$, and $\mu(\cdot)$, $\varphi(\cdot)$ denotes two dimensional transformation operation functions.

\noindent
{\bf Hypergraph Convolution.} We use transposition methods to enable the construction of hypergraph structures with additional learning to the feature of the hyperedges~\cite{DBLP:conf/nips/JoBLKKH21}. We use the following hypergraph convolution to capture important high-order feature information~\cite{DBLP:journals/tip/HaoLGJY21}:
\begin{align}
HC^{T}(H_{i})=X_{i}\cdot  (\widetilde{H_{i}}+\widetilde{H^{T} _{i}})\cdot  \Theta _{i},
\end{align}
where $\Theta$ represents the learning weights, which are implemented in this paper by different convolution methods, $H_{i}$ is a hypergraph of different types of constructions, $\widetilde{H_{i}}=D_{v}^{-1/2} H_{i}WD_{e}^{-1}H_{i}^{T} D_{v}^{-1/2}$~\cite{DBLP:conf/aaai/FengYZJG19}, $D_{v},D_{e}$ are the diagonal matrices of hyperedges and node degrees respectively, $W$ is the diagonal matrix, initialized to the unit matrix, $X_{i}\in R^{C_{in}\times T\times V}$ is the input feature,.

In addition, $HC(H_{i})$ denotes the hyperedge feature is not considered: 
\begin{align}
HC(H_{i})=X_{i}\cdot  \widetilde{H_{i}}\cdot  \Theta _{i}
\end{align}
Finally, we obtain the following features: $HC^{T}(H_{N})$, $HC^{T}(H_{T})$, $HC(A)$, $HC(H_{ST})$, $HC(H_{TS})$, denoted by $A^{'}-E^{'}$, respectively. The implementation process is shown in Fig.~\ref{fig02}.
\subsection{High-order Information Fusion (HIF)}
\label{CC}


With the multiple high-order information obtained from convolution operations, we constructed the HIF module for parallel modeling. The implementation is shown in Fig.~\ref{fig02}. First, we use temporal features to refine spatial features to obtain high-order joint information. Then through spatial-temporal aggregation in the channel dimension and channel concatenating operations to capture the properties between features. The final output is  $F_{out}\in R^{C_{out}\times T\times V}$.
\begin{figure}[t] 
\centering 
\includegraphics[width=0.9\linewidth,height = 0.9\linewidth]{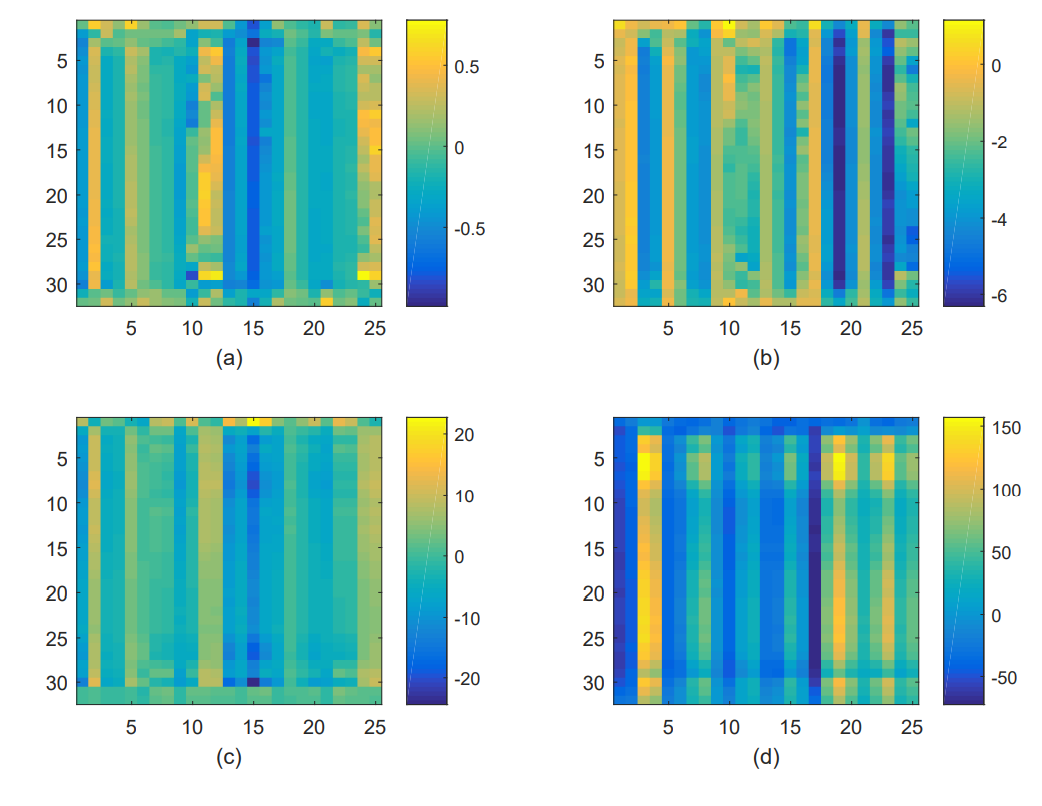} 
\caption{(a), (c) Spatial-temporal features of the TPH and k-means spatial topology. (b), (d) Spatial-temporal features of the TPH and k-NN spatial topology.} 
\label{fig11} 
\end{figure}

Specifically, to avoid the loss of the original topology information, we use the original natural joint topology as an auxiliary learning component. It is combined with spatial hypergraph features to learn more connected joint features of the action. As indicated below:
\begin{align}
Y_{2}=(HC^{T}(H_{N})+HC(A))/\beta,
\end{align}
where $\beta$ is the learning weights.




\subsection{Model Architecture}
\label{DD}
We use a multi-scale convolution, TF, with an attention mechanism~\cite{wang2022skeleton,2021Channel} to extract information at different scales. As expressed in the following equation:
\begin{align}
Z_{out} =TF\left ( F_{out1}+F_{out2}+F_{out3}      \right ).
\end{align}

$Z_{out}\in R^{C_{out}\times T\times V}$ is the output of a block of dynamic spatial-temporal hypergraph convolutional network. The network consists of ten blocks. The difference in $F_{out1}$, $F_{out2}$, $F_{out3}$ is composed of three different spatial hypergraph topologies, respectively.

\section{Experiments}

\subsection{Datasets}
\label{11}
\noindent
{\bf Northwestern-UCLA.} The dataset~\cite{DBLP:journals/corr/WangNXWZ14} is extracted from 1494 videos with ten types of actions. The training set comes from two Kinect cameras and the test set from remaining cameras.

\noindent
{\bf NTU RGB+D}. NTU RGB+D~\cite{2016NTU}  is a large and widely used action recognition dataset obtained from three different views using the Kinect v2 sensor and contains 60 action classes. Evaluation benchmarks: Cross-Subject (X-Sub) and Cross-View (X-View).

\noindent
{\bf NTU RGB+D 120.} NTU RGB+D 120~\cite{2020NTU} is an enhanced version of the NTU RGB+D dataset, with the number of action categories increased to 120. Evaluated using two division criteria: Cross-Subject (X-Sub) and Cross-Setup (X-Setup).
\subsection{Implementation Details}
\label{22}
Our experiments use cross entropy as the loss function and use three Tesla M40 GPUs. In addition, we use dual correlation channel data to aid fuse learning, which is obtained by subtracting two channels. The experiments use the stochastic gradient descent (SGD). The weight decay is set to 0.0004 and the Nesterov momentum is 0.9. The train contains 90 epochs, the first five epochs use a warm-up strategy, and we set cosine annealing for decay~\cite{loshchilov2016sgdr}. For the NTU-RGB+D~\cite{2016NTU} and NTU-RGB+D 120 datasets~\cite{2020NTU}, the batch size is set to 64. For the Northwestern-UCLA dataset~\cite{DBLP:journals/corr/WangNXWZ14}, we set the batch size to 16. 
\begin{table}[t]
 \centering
\caption{The ablation study of components.}

\setlength{\tabcolsep}{1.5mm}
\label{tab:34}       
\begin{tabular}{lllll}
\hline\noalign{\smallskip}
 Methods &FLOPs&Param. & Acc($\%$)\\
\noalign{\smallskip}\hline\noalign{\smallskip}
DST-HCN w/o C,H,G&3.05G&2.71M&89.5\\
DST-HCN w/o C,H&3.50G&2.93M&90.1 ($\uparrow$ 0.6$\%$)\\
DST-HCN w/o C&3.50G&2.93M&90.4 ($\uparrow$ 0.9$\%$)&\\
DST-HCN&3.50G&2.93M&90.9 ($\uparrow$ 1.4$\%$)\\
\noalign{\smallskip}\hline
\end{tabular}
\end{table}
\subsection{Ablation Study}
Unless otherwise stated, we analyze our method on the X-sub benchmark of the NTU RGB+D dataset.
\begin{table*}[ht]
 \centering
\caption{Comparison of recognition accuracy between hypergraph and graph network in confusing actions. The experiments are conducted on the X-sub benchmark of the NTU RGB+D 120 dataset. Where wield knife representative wield knife towards other person, hit other representative hit other person with something.}
\label{confu}       
\setlength{\tabcolsep}{1.5mm}{
\begin{tabular}{llllll}
\hline\noalign{\smallskip}
Action &Ens Acc($\%$)&Similar Action &Ours Acc($\%$)&Acc($\%$)$\uparrow$&Similar Action\\
\noalign{\smallskip}\hline\noalign{\smallskip}
writing&48.90&typing on a keyboard&60.66&11.76&typing on a keyboard\\
wield knife&62.50&hit other &65.80&3.30&hit other \\
blow nose&59.65&yawn&64.52&4.87&yawn\\
fold paper &62.78&ball up paper &66.96&4.18&counting money\\
yawn &67.65&hush (quite)&70.26&2.61&hush (quite)\\
snapping fingers &65.51&shake fist&71.43&5.92&make victory sign\\
sneeze/cough&71.74&touch head (headache) &75.72&3.98&touch head (headache) \\
apply cream on hand back&72.30& open bottle &75.26&2.96& rub two hands together \\

\noalign{\smallskip}\hline
\end{tabular}}
\end{table*}

\begin{table*}[ht]
	\centering
	\caption{ Classification accuracy comparison with state-of-the-art methods on the three benchmarks.}
	\setlength{\tabcolsep}{3mm}
	\label{1}
	\begin{tabular}{c|c|c|c|c|c|c}
	\hline
	\multirow{2}{*}{Type}&
	\multirow{2}{*}{Method} &
	\multicolumn{2}{c|}{NTU RGB+D} &
    \multicolumn{2}{c|}{NTU RGB+D 120} &
	\multicolumn{1}{c}{NW-UCLA} \\
    \cline{3-7}
	& &X-Sub ($\%$) &X-View  ($\%$)&X-Sub  ($\%$)&X-Set  ($\%$) &-  ($\%$)\\
	\hline
	    \multirow{8}{*}{GCN,CNN}&
		ST-GCN~\cite{DBLP:conf/aaai/YanXL18}&81.5&88.3&70.7&73.2&-\\
		&2s-AGCN~\cite{2018Two}&88.5&95.1&82.5&84.2&-\\
		&Shift-GCN~\cite{DBLP:conf/cvpr/Cheng0HC0L20}&90.7&96.5&85.9&87.6&94.6\\
		&SGN~\cite{2020Semantics}&89.0&94.5&79.2&81.5&92.5\\
		&CTR-GCN~\cite{2021Channel}&{\bf 92.4}& {\bf96.8}&{\bf88.9}&\underline{90.6}&\underline{96.5}\\
		&Ta-CNN~\cite{xu2022topology}&90.4&94.8&85.4&86.8&96.1\\
		&EfficientGCN-B4~\cite{2021Constructing}&92.1&96.1&88.7&88.9&-\\
		&AGE-Ens~\cite{DBLP:journals/corr/abs-2105-01563}&92.1&96.1&88.2&89.2&-\\
	\hline
	    \multirow{4}{*}{HypGCN}&
	    Hyper-GNN~\cite{DBLP:journals/tip/HaoLGJY21} &89.5&95.7&-&-&-\\
	    &DHGCN~\cite{wei2021dynamic}&90.7&96.0&86.0&87.9&-\\
	    &Selective-HCN~\cite{zhu2022selective}&90.8&\underline{96.6}&-&-&-\\
	 \cline{2-7}
	    &DST-HCN(Ours)&\underline{92.3}&{\bf 96.8}& \underline{88.8}&{\bf 90.7}&{\bf 96.6}\\
	    \hline
	\end{tabular}
\end{table*}
\noindent
{\bf Visualization experiments.} We demonstrate the spatial-temporal features of the two topologies of an action sample "putting on glasses" in Fig.~\ref{fig11}.  (a), (b) are the output of the HIF module "$C^{'}$" in Fig.~\ref{fig02}. (c), (d) are the output of HIF module "$F_{out}$". The coordinates are time and joints respectively.


In Fig.~\ref{fig11} (a), (b), We observe that different combinations of hypergraphs is able to learn different spatial-temporal information. In addition, (a) focuses more on information from the elbow joint and (b) focuses on information features of the hand, suggesting that these joints are more discriminatory for overall. In Fig.~\ref{fig11} (c), (d), the further changes in joint features occurred and the joints are differentiated in temporal aspects. It is shown that the channel learning method can effectively perform information fusion.

\noindent
{\bf Ablation experiments of different components.} We experimentally validate the importance of the components, where w/o G, H, and C denote the lack of the original natural topology, the original topology as a hypergraph structure, and the lack of cosine annealing, respectively. Table~\ref{tab:34} shows that the model without cosine annealing has a performance drop of 0.5$\%$. The result drops by 0.6$\%$ without using the original graph structure, which shows that the unnatural joint connection needs to learn certain a prior knowledge of the natural topology. In addition, the result of using the original topology as the hypergraph structure by a margin of 0.3$\%$, reflecting the hypergraph structure's advantage and is more beneficial for action recognition.

\noindent
{\bf The comparison of confusing actions.} The DST-HCN and the fine-grained recognition graph convolution method~\cite{DBLP:journals/corr/abs-2105-01563} were compared in terms of fine-grained action recognition performance. As shown in Table~\ref{confu}, the recognition performance of all relevant actions is improved. Especially in the writing and snapping fingers actions by a margin of 11.76$\%$ and 5.92$\%$, respectively. This indicates that our model can handle unnaturally dependent spatial-temporal variation features well.
\subsection{Comparisons With the State-of-the-Art Methods}
\label{33}
This section compares our dynamic spatial-temporal hypergraph neural network with state-of-the-art methods. For a fair comparison, we use four streams of fusion results, considering joint, bone, and motion data stream forms~\cite{2021Channel, DBLP:conf/cvpr/Cheng0HC0L20,2018Two}. We use different fusion parameters~\cite{wang2022skeleton} for different data to accommodate the diversity of varying stream forms.

The results are shown in Table~\ref{1}. It can be seen that our model exhibits comparable or better performance than the state-of-the-art approach for the three datasets. 
Compared to the hypergraph methods, our proposed method has reached state-of-the-art accuracy in hypergraphs and by a margin of 1.5$\%$ and 2.8$\%$ on X-Sub, which fully illustrates the significant advantage of our method.

\section{Conclusion}
This paper proposes a dynamic spatial-temporal hypergraph neural network that contains dynamic spatial-temporal hypergraph convolution and high-order information fusion modules. The method presents a variety of hypergraph structures, such as a time-point hypergraph. The method simultaneously implements parallel spatial-temporal modeling. Finally, we validate the model on the NTU RGB+D, NTU RGB+D 120, and NW-UCLA datasets.
%

\bibliographystyle{IEEEbib}
\bibliography{icme2023template}

\end{document}